\documentclass[journal]{IEEEtran}
\usepackage{amsmath,amsfonts}
\usepackage{algorithmic}
\usepackage{algorithm}
\usepackage{array}
\usepackage[caption=false,font=normalsize,labelfont=sf,textfont=sf]{subfig}
\usepackage{textcomp}
\usepackage{stfloats}
\usepackage{url}
\usepackage{verbatim}
\usepackage{graphicx}
\usepackage{cite}
\usepackage{multirow}
\begin{document}

\title{Lightweight Operations for Visual Speech Recognition}

\author{Iason Ioannis Panagos,~\IEEEmembership{Member,~IEEE},
Giorgos Sfikas,~\IEEEmembership{Member,~IEEE},
Christophoros Nikou,~\IEEEmembership{Senior Member,~IEEE}
\thanks{I. Panagos and C. Nikou are with the Department of Computer Science \& Engineering, University of Ioannina, Ioannina, Greece.}
\thanks{G. Sfikas is with the Department of Surveying \& Geoinformatics Engineering, University of West Attica, Athens, Greece.}}




\markboth{Journal of \LaTeX\ Class Files,~Vol.~14, No.~8, August~2021}%
{Panagos \MakeLowercase{\textit{et al.}}: Lightweight Operations for Visual Speech Recognition}


\maketitle

\begin{abstract}
Visual speech recognition (VSR), which decodes spoken words from video data, offers significant benefits, particularly when audio is unavailable.
However, the high dimensionality of video data leads to prohibitive computational costs that demand powerful hardware, limiting VSR deployment on resource-constrained devices.
This work addresses this limitation by developing lightweight VSR architectures.
Leveraging efficient operation design paradigms, we create compact yet powerful models with reduced resource requirements and minimal accuracy loss.
We train and evaluate our models on a large-scale public dataset for recognition of words from video sequences, demonstrating their effectiveness for practical applications.
We also conduct an extensive array of ablative experiments to thoroughly analyze the size and complexity of each model.
Code and trained models will be made publicly available.
\end{abstract}


\section{Introduction}\label{sec:introduction}

\textit{Visual Speech Recognition} (VSR) is a computer vision problem which aims to decode speech present in visual media of one or more speakers without the presence of sound.
Alongside \textit{Audio Speech Recognition} (ASR) which is its counterpart regarding audio-only media, such as recordings, it is a subset of the more general problem of \textit{Speech Recognition} (SR).

Applications of SR in everyday life can be found in several domains, notably in the medical assistance, where such a system can be utilized to provide assistance to patients that are speech-impaired or otherwise have communication difficulties.
In the same spirit, various instances of existing accessibility platforms (e.g., mobile devices and interfaces) can benefit from the addition of a SR system, improving the everyday lives of many individuals.

Automatic SR systems have been adopted by the entertainment industry as well, where they have been employed to automatically generate captions of video segments, short clips and even entire full-length movies, or to produce transcriptions for older, silent films, by relying only the video stream.
A striking example is the "automatic captions" feature provided by various video hosting platforms, such as \textit{YouTube}, where an ASR unit generates captions for some videos using only the audio track.
Employing such systems for the purpose of creating media transcriptions can save significant time and effort, when compared to using human lip readers or speech transcribers, as automatic systems have now surpassed humans for this task in both speed and accuracy, streamlining the overall process by reducing the associated costs.

Recognizing speech units using only an audio signal is considered to be a less-demanding task, due to the lower dimensions of the audio stream (one-dimensional sequence), when compared with the spatio-temporal aspects associated with a video (three-dimensional sequence).
Additionally, the amount and variety (e.g., languages) of publicly available audio data far surpasses that of video, allowing for a wider adoption and deployment of ASR systems.
As a result of these factors, VSR models have been adapted to more specialized use-cases or in a secondary, auxiliary capacity to assist existing ASR units.
However, in cases where applications of ASR are rendered ineffective by a significant amount of noise (e.g., crowded environments, multiple speakers) or in media where the audio track does not exist, such as in silent video recordings and films, employing a VSR system remains the only option.

In contrast to ASR, relying on visual cues to recognize speech is a more challenging process which involves more sophisticated and powerful architectures in order to produce meaningful results.
An important distinction between the two modalities (audio and video) lies in the form of data to be processed, since a video contains spatio-temporal data of higher dimensionality compared to an audio stream and is therefore more demanding on computing resources.
In addition, visual ambiguities between words that are produced from similar or identical mouth movements can cause erroneous results.
A typical example includes the plural version of a word where the added suffix is hard to distinguish using only the visual information, while the added audio cue at the end of the word greatly contributes in successfully predicting the correct word.
Distinguishing between such words demands powerful models with sufficient representation capabilities and as a result, VSR-specific systems rely on deep, large-sized models in terms of parameters in order achieve high performance.
These networks also suffer from higher latency and prohibitive computational costs, factors which hinder their applicability in practical scenarios or applications where speed of operation is critical (e.g., embedded devices).

A few of the aforementioned use cases of SR which are met in real-life scenarios typically demand on-line data processing and acceptable performance (e.g., low runtime) in order to be useful for the end-user.
In this article, our goal is developing lightweight and compact architectures for visual speech recognition of words in order to enable such applications.
To that end, we design deep neural networks by utilizing cost-effective network components that take advantage of operations with low computational overhead.
Our models benefit from low sizes in terms of required parameters as well as reduced computational complexity, making them ideal for various practical applications.
We conduct an extensive experimental analysis which showcases that our models feature greatly reduced hardware demands, without compromising their accuracy.

In summary, our contributions are the following:
\begin{itemize}
	\item We employ Ghost modules in a unified VSR architecture by replacing the standard convolutions in its components (visual feature extractor and sequence model) in order to reduce its overall computational overhead.
		  Using Ghost modules, we further reduce the running costs of two established temporal convolution architectures that are used for sequence modeling, resulting in models that are even more lightweight than their standard versions and achieve comparable accuracy.
		  The final architecture still performs very competitively compared to the original, while being less demanding in resources, measured in terms of model parameters and computational overhead.
	\item We also design a general temporal block architecture, named \textit{Partial Temporal Block}, that splits the input volume in two parts and applies separate operations in each part.
		  Using this component as a building block, we follow three methods from the literature and develop ultra-lightweight temporal convolution networks aimed at applications with very low power.
	\item We perform extensive experiments on the largest publicly-available dataset for English words and our results showcase strong visual speech recognition performance.
		  Simultaneously, our proposed models are practical in terms of hardware demands, as showcased in a detailed ablative analysis, allowing for several applications by devices with varying computation capabilities.
\end{itemize}

\section{Related Work}\label{sec:related}

The task of visual speech recognition has been under active research for several decades.
In order to tackle the problem, the paradigm typically followed by the literature splits it into simpler sub-problems and involves a series of steps.
Initially, a spatial processing step aims to extract high-dimensional feature representations from the input.
Subsequently, a sequential modeling step interprets the temporal inter-relations between the feature representations of each time-step of the sequence.
Finally, a classification step aims to correctly predict the spoken word depicted in the frames.

Earlier works commonly employed simple image transform techniques such as \textit{Principal Component Analysis} or \textit{Discrete Cosine Transform} for visual feature extraction from lip area images, while \textit{Support Vector Machines} or \textit{Hidden Markov Models} were used as classifiers \cite{potamianos2003recent}.
Their vocabularies consisted of few words or single digits and as a result their applicability in real-life scenarios was rather limited.
Furthermore, the available hardware at the time was insufficient to handle the non-trivial computational overhead, constraining the deployment and application of such methods.
With recent progress in both of these domains, research efforts have been increased and powerful models achieving impressive results have been developed.

Due to the remarkable advances in machine learning research of the last decade, the commonly-followed approach employs deep neural networks for the two initial steps, and a single densely-connected layer for the latter, since these architectures have demonstrated high performance on such tasks.
For visual feature extraction, \textit{Convolutional Neural Networks} (CNN) have been established as the primary model of choice due to their ability to extract strong representations when trained on sufficient amounts of data.
For the second step, the models used for sequential processing of the extracted features have been predominantly based on recurrent architectures, such as \textit{Long Short-Term Networks} (LSTM) \cite{chung2018learning, stafylakis2017deep, stafylakis2018pushing} and \textit{Gated Recurrent Units} (GRU) \cite{xu2020watch, miao2020part, liu2021robust}.
The RNNs that are employed are typically set as \textit{bi-directional} where they also process the reverse of the sequence, and subsequently their outputs are fused by concatenation.

More recently, variants of \textit{Temporal Convolution Networks} (TCN) \cite{lea2017temporal} have been proposed as alternatives to recurrent neural networks for sequential tasks, offering higher performance.
Such architectures are gradually replacing recurrent ones due to their favorable characteristics regarding training stability and model simplicity \cite{bai2018empirical}.
An approach that utilizes several convolutions with different kernel sizes in each block of the standard TCN architecture is introduced in \cite{martinez2020lipreading}.
This model leverages the different effective receptive fields of the convolution kernels to increase its representation capabilities by incorporating more features across the time domain and has been adopted by several recent works (e.g., \cite{sheng2022importance, tian2022lipreading, peng2022lip}).
A more complex model building upon the multi-kernel approach is proposed in \cite{ma2021lip}, where dense connections are added in the architecture.
In this way, more features are utilized per stage, increasing the model's depth and expressiveness at the cost of its size and required calculations.

For single word recognition, while a few works utilize only the visual stream (e.g., \cite{chung2018learning, martinez2020lipreading}), others propose methods that utilize both streams in a complementary fashion to boost performance (e.g., \cite{xu2020watch, liu2021robust}). 
Typically, modality fusion mechanisms of various complexities are employed to seamlessly integrate information from the video and audio streams.
For instance, while a simple concatenation operation is used in \cite{xu2020watch}, \cite{liu2021robust} propose a hybrid fusion network that utilizes features from both audio and video modalities with a decision fusion mechanism to predict the final word.

The majority of published works focuses on improving word recognition accuracy without considering the associated computational overhead that is a consequence of using sizable models that integrate several, oftentimes complex, components in their architectures.
Consequently, the proposed models cannot be utilized in a resource-restricted environment due to the significant hardware requirements.
In comparison, research aimed at lowering model complexity and improving efficiency has not received as much attention and remains at an early stage, with fewer works appearing recently.

Models intended for applications by low-power hardware such as mobile devices were proposed in \cite{shrivastava2019mobivsr}, where the authors design low cost networks by following lightweight convolutional neural network principles.
More specifically, a compact spatio-temporal module is introduced in order to improve the performance of video recognition.
It is combined with an architecture for visual feature extraction that is designed to be efficient by combining blocks with residual connections and depth-wise convolutions.
A scaling factor controls the balance between performance and accuracy by adjusting the network size.
To further improve the running speed and lower memory-intensity of the proposed models, a simple sequential network based on temporal convolutions is adopted for modeling the extracted features of the entire sequence.

In order to reduce the computational complexity of the entire visual speech recognition process, \cite{ma2021towards} propose changing each component used for feature extraction and sequence modeling.
Using the model from \cite{martinez2020lipreading} as a baseline, a lightweight convolutional neural network is used for visual feature extraction reducing the hardware requirements in terms of parameters and processing operations.
Then, to further reduce the overall model's overhead, a lightweight block is introduced to the TCN architecture, replacing its standard operations with the commonplace \textit{depthwise-separable} design paradigm of lightweight CNNs (e.g., \cite{chollet2017xception, howard2017mobilenets, zhang2018shufflenet}).
Finally, to recover some of the accuracy lost due to the drop in network capacity, a form of knowledge distillation is used to train the models.

A large study benchmarking several deep learning architectures for extraction as well as sequence modeling was performed in \cite{arakane2023efficient}.
The authors train and evaluate an extensive selection of recently-proposed models on a variety of publicly available datasets for both English and Chinese languages.
The architectures used in their experiments cover a wide range of networks for feature extraction as well as sequence modeling, such as convolutional, vision transformers and temporal convolution networks.

A more recent approach \cite{panagos2022compressing, panagos2024visual} involves applying a parameter sharing technique to compress the components of VSR systems leading to more compact models without compromising accuracy.
More specifically, the convolutional layers employed by both components in the VSR pipeline (i.e., feature extractor and sequential model) are replaced equivalent layers following a formulation that exploits a sum of Kronecker products to enable parameter sharing, greatly reducing the required size of each layer.
The models achieve significant reductions in size and parameters for a minor performance penalty, which becomes more pronounced for higher rates of compression.

\section{Method}\label{sec:method}

\subsection{Proposed Model}\label{sub:proposed}

\begin{figure}[ht]
	\centering
	\includegraphics[width=\linewidth]{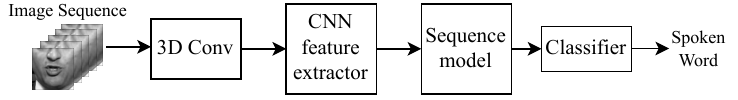}
	\caption{Overview of the architecture used for visual speech recognition. We experiment with several feature extractors as well as our proposed lightweight sequence models. The Softmax function is used as the classifier. The overall system outputs a spoken word.}
	\label{fig:model}
\end{figure}

The architecture of our proposed model follows the two-step design paradigm outlined in Section~\ref{sec:related}, and its general structure is depicted in Figure~\ref{fig:model}.
We experiment with efficient components to design lightweight speech recognition models with affordable computation demands.
For both feature extraction and sequence modeling, we employ Ghost modules (Section~\ref{sub:ghost}), greatly reducing network overhead, while we also propose a Partial Temporal Block (Section~\ref{sub:partial}) to develop ultra-lightweight TCN-based architectures suitable for scenarios with very-low-powered hardware.
Using these components, our models can be deployed in several applications due to their low resource requirements.

\subsection{Ghost Module}\label{sub:ghost}

Ghost modules were proposed in \cite{han2020ghostnet} as a component that takes advantage of ``cheap operations" to reduce its computation cost compared to a typical convolution layer.
A ghost module achieves low computational overhead in two steps.
First, a regular $1 \times 1$ convolution generates a set of feature maps from the input.
A fixed ratio determines the number of channels in the generated feature maps, controlling the resource savings of the component.
Typically, the ratio is set to $0.5$ meaning that channels in the produced feature maps equal half of the input volume's channels.

A ``cheap operation" uses these intermediate feature maps to produce an additional set with the same channel size.
The role of the cheap operation can be undertaken by any lightweight function, in the Ghost module, a depth-wise convolution with a kernel size of $3 \times 3$ is used.
This convolution operates on each filter and processes the spatial information it contains, while preserving the amount of channels.
Finally, the two distinct feature maps are concatenated along the channel dimension, meaning that the output volume matches the input's channels.

Compared to the standard convolution operation, this formulation reduces the total amount of computation required since the initial $1 \times 1$ convolution generates feature map with fewer channels, and the depth-wise operation which is much cheaper computationally, is also applied on this volume, rather than the whole input.
By preserving the original output size of a convolution layer, a Ghost module can act as a drop-in replacement for that layer to reduce computational overhead in a network architecture.
The operations of the Ghost module can be summarized as:
\begin{align}
	X_1 = ReLU(BatchNorm(Conv_{1\times1}(X))) \nonumber \\
	X_2 = ReLU(BatchNorm(DWConv_{3\times3}(X_1))) \nonumber \\
	Out = concatenate([X_1, X_2])\,, \label{eq:ghost}
\end{align}

\noindent where $X$ refers to the input volume and \(DWConv\) to the depth-wise convolution.

A drawback related to the representation capabilities of the Ghost module arises from the fact that the initial $1 \times 1$ convolution reduces the feature map channel dimensionality (to half) in order to keep the costs of the module low.
Subsequently, the second ($3 \times 3$ depth-wise) convolution operates on a sub-set of the input feature map and might miss some spatial relationships that would otherwise be captured by operating on the full input volume.
Since half of the final feature map in the output of the Ghost module is produced from the $1 \times 1$ convolution without any spatial interaction between the pixels, the performance of the module is constrained.
To alleviate this weakness, the authors of \cite{tang2022ghostnetv2} propose an enhancement called \textit{DFC attention} which aims to exploit long-range spatial information, augmenting the Ghost module's intermediate features with richer representations that were lost by the original design.

DFC attention utilizes two fully-connected layers which are applied to the input features in a sequential manner, spanning both the vertical and horizontal directions and aggregating the features in each direction.
By operating on the different directions separately instead of on a square area, the computational complexity of the attention module is kept low.
Initially, the input feature map is spatially down-sampled both vertically and horizontally with a pooling operation which shrinks the spatial dimensions by half.
Since the subsequent layers operate on feature maps of smaller size, the required computations are reduced significantly.
Then, the fully-connected layers are applied in a sequential manner, first the vertical (column-wise) layer, followed by horizontal (row-wise) layer.
Finally, the produced feature map passes through a non-linear activation function to scale its values in the $(0, 1)$ range, producing an attention map and an up-sampling operation restores the original spatial dimensions.
The DFC module is implemented with a pooling operation that averages the values, while the non-linearity at the end is handled with a Sigmoid function. 
The following equations show the DFC module's operations on an input volume $X$:

\begin{align}
	X_1 = AveragePool(X) \nonumber \\
	X_2 = BatchNorm(Conv_{1\times1}(X_1)) \nonumber \\
	X_3 = BatchNorm(Conv_{1\times5}(X_2)) \nonumber \\
	X_4 = BatchNorm(Conv_{5\times1}(X_3)) \nonumber \\
	X_5 = Sigmoid(X_4)\,. \label{eq:dfc}
\end{align}

Adding DFC attention to the Ghost module incurs an increase in parameter size due to the additional convolutions but only a slightly higher computation cost in FLOPs.
However, in practice, when tested on mobile devices, it achieves better performance at the same latency \cite{tang2022ghostnetv2}.

\begin{figure*}[!th]
	\centering
	\subfloat[]{\includegraphics[width=0.20\textwidth]{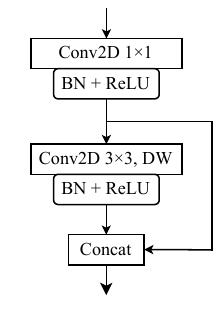}
		\label{fig:ghost_block}}
	\hfil
	\subfloat[]{\includegraphics[width=0.20\textwidth]{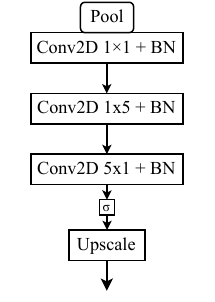}
		\label{fig:dfc_attention}}
	\hfil
	\subfloat[]{\includegraphics[width=0.20\textwidth]{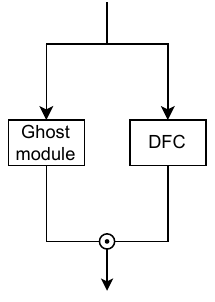}
	\label{fig:ghostv2_block}}
	\caption{Ghost modules. BN indicates the \textit{Batch Normalization} operation, ReLU indicates the \textit{Rectified Linear Unit} function, DW refers to the \textit{depth-wise convolution}, while $\sigma$ is the logistic \textit{Sigmoid} and $\odot$ is the element-multiplication sign. (a) Original Ghost Module \cite{han2020ghostnet}. (b) DFC attention \cite{tang2022ghostnetv2}. (c) Ghost Module with DFC attention.}
	\label{fig:ghost_modules}
\end{figure*}

\subsection{Partial Temporal Block}\label{sub:partial}

Reducing the size of the input feature map and operating on the resulting tensor is an effective approach to reduce the computational overhead of a network component, that has been followed by several works (e.g., \cite{chollet2017xception, howard2017mobilenets}).
Within a network block, using the initial layer to reduce the channel dimension of an input volume, and applying the subsequent layers in the produced, smaller output, allows controlling the amount of calculations and enables the development of lightweight network components with low operating costs.
An additional operation, typically the final one in a block, restores the channel dimension to match that of the input, usually in order to facilitate a residual connection.
This design is commonly known as a \textit{bottleneck}, since the intermediate feature maps have a lower number of channels.

A similar approach \cite{ma2018shufflenet, wang2020cspnet, chen2023run} splits the input feature map across the channel dimension in two parts according to a fixed ratio, and applies two separate branches, one in each part.
The operations in either branch can have any form, for instance, in \cite{chen2023run} a regular convolution followed by two point-wise layers is applied on one branch, while the second branch leaves the input unchanged.
To form the output, the results of each separate branch are merged along the channel dimension via concatenation.

\begin{figure*}[!th]
	\centering
	\subfloat[]{\includegraphics[width=0.275\textwidth]{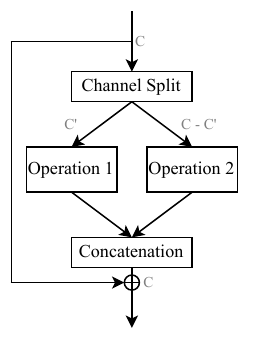}
		\label{fig:partial_block}}
	\hfil
	\subfloat[]{\includegraphics[width=0.20\textwidth]{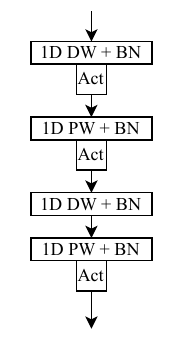}
		\label{fig:shufflenet_block}}
	\hfil
	\subfloat[]{\includegraphics[width=0.20\textwidth]{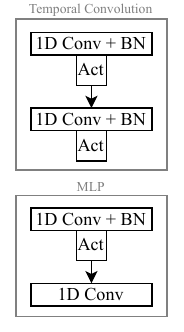}
		\label{fig:fasternet_block}}
	\caption{Block designs used in the proposed \textit{Partial Temporal Block}. (a) The block architecture. ``C'' represents the amount of channels of the input volume to each component. (b) \textit{ShuffleNet} \cite{ma2018shufflenet} block architecture. (c) FasterNet \cite{chen2023run} block components. ``DW'' and ``PW'' indicate \textit{depth-wise}, and \textit{point-wise} convolutions. ``BN'' is the \textit{Batch Normalization} layer and ``Act'' can be any activation function (e.g., ReLU).}
	\label{fig:blocks}
\end{figure*}

Inspired by the practicality and results of methods following this paradigm (e.g., \cite{ma2018shufflenet, chen2023run}), we design the Partial Temporal Block, which follows the same principle.
Our block allows for a wide network design flexibility as it can be tailored to each specific application constraints (e.g., hardware capabilities, dataset availability and size), and can even be part of a search space, in order to obtain the most optimal setup, depending on the problem.
For an input volume $X$, the operations of the partial block can be summarized as:

\begin{align}
	X_1, X_2 = Channel\_split(X) \nonumber \\
	X_3 = F(X_1) \nonumber \\
	X_4 = G(X_2) \nonumber \\
	X_{c} = concatenate([X_3, X_4]) \nonumber \\
	X_{out} = X_{c} + X \,, \label{eq:partial}
\end{align}

\noindent where the channel split operation divides the input in two parts along the channel dimension according to a fixed ratio, $F$ and $G$ can be any type of operation, including sequences of layers, and the final concatenation merges the output of each branch in the channel dimension.
A skip connection with the input is also added to facilitate easier training of deep architectures.
The general block architecture is depicted in Figure~\ref{fig:partial_block}(a).

As a baseline, we employ the standard Temporal Convolution layer \cite{bai2018empirical} as the core of our block in one branch.
This layer uses a sequence of 1D causal convolutions with batch normalization and non-linear activation functions, repeated twice.
The other branch uses no operations, this way the computational overhead of the block is greatly reduced.
The overall architecture is comprised of four stages, where each stage is one Partial Temporal Block with increasing dilation rate for the non-point-wise convolutions.
This way, the entire network is very lightweight in terms of hardware requirements (see Section~\ref{sub:param_analysis}).

Furthermore, following the designs of \cite{ma2018shufflenet} and \cite{chen2023run}, we construct two other highly efficient (also four-stage) TCN-based networks that require few parameters and have very low computational overhead in terms of FLOPs.
Their operations as used within our proposed block are depicted in Figure~\ref{fig:blocks} (b) and (c).
We note that, for the \textit{ShuffleNet} \cite{ma2018shufflenet} block design, a channel mixing operation is added at the very end (after concatenation and addition), while for the \textit{FasterNet} \cite{chen2023run} block design, the MLP network is applied after concatenation of the branches and before adding the input via the skip connection.

\section{Experiments}\label{sec:experiments}

\subsection{Dataset \& Preprocessing}\label{sub:dataset}

This work uses the \textit{Lip Reading in the Wild\footnote{\url{https://www.robots.ox.ac.uk/~vgg/data/lip_reading/lrw1.html}}} (LRW) dataset \cite{chung2017lip} for model training and evaluation.
LRW features a rich vocabulary of $500$ distinct words spoken from a variety of more than $1000$ speakers in short segments recorded from public television programs and therefore exhibits variations in the backgrounds as well as the speakers.
While the scene background generally varies depending on the program, lighting conditions are adequate and the speakers are clearly visible.
Multiple angles of persons speaking are also present, adding to the complexity of the dataset. 

The LRW dataset is split into three subsets (train, validation and test) without overlapping segments, and each sequence spans $29$ frames at a fixed frame rate of $25$ FPS.
A single word utterance occurs at the middle of each video sequence.
The total length of the dataset's segments amounts to $173$ hours.
Details about the dataset splits are shown in Table~\ref{tab:lrw}.

\begin{table}[th]
	\caption{Dataset split details for LRW \cite{chung2017lip}.}
	\label{tab:lrw}
	\centering
	\begin{tabular}{l r c r}
		\hline
		\textbf{Split} & 
		\multicolumn{1}{c}{\textbf{Samples}} & 
		\multicolumn{1}{c}{\textbf{Sequences/word}} & \multicolumn{1}{c}{\textbf{Hours}} \\
		\hline
		Train & $488.766$ & $800-1000$ & $157.50$ \\
		Validation & $25.000$ & $50$ & $8.05$ \\
		Test & $25.000$ & $50$ & $8.05$ \\
		\hline
	\end{tabular}
\end{table}

To prepare the raw data for training, we employ a simple procedure that is typically used by previous works in the literature (e.g., \cite{martinez2020lipreading, feng2021efficient}).
First, landmarks are computed using a face alignment network after the face of the speaker has been detected in each frame.
Then, to keep images uniform throughout the set, size and rotation variations are removed by using a mean face shape and the mouth regions of interest are cropped with a $96 \times 96$ bounding box.
Finally, normalization by mean and standard deviation and conversion to gray scale is is applied.

\subsection{Training Setup}\label{sub:training}

Networks and experiments are implemented using the PyTorch framework\footnote{\url{https://pytorch.org}}.
All models are trained from randomly initialized weights on the LRW training set.
An initial learning rate of $0.01$ with a cosine annealing schedule is used.
To prevent over-fitting, weights are decayed by $0.01$ and dropout on the TCN layers for all models is set to $0.2$.
We train for a total of 80 epochs with Stochastic Gradient Descent, using a batch size of $32$, without any warming up period.
During training, spatial cropping flipping are randomly applied, as well as MixUp \cite{zhang2018mixup} and variable length augmentation \cite{martinez2020lipreading}.
After each epoch, the model is validated and the best performing checkpoints are saved.

\subsection{Results \& Discussion}\label{sub:results}

Our proposed models are evaluated in the LRW test set and in Table~\ref{tab:results} we provide a comparison with other lightweight models from the literature.
The metric used to evaluate the methods is word accuracy, measured as a percentage of correct word predictions.
We also include size and model complexity measurements, more specifically, the amounts of total network parameters and Floating Point OPerations (FLOPs), as these values are useful when gauging the overall practicality of the methods when considering several applications.
More detailed, per-model overviews are provided in Tables~\ref{tab:params_flops_ghost} and~\ref{tab:parameter_analysis}.
All measurements are obtained using \textit{torchinfo}\footnote{\url{https://github.com/TylerYep/torchinfo}}.

\begin{table*}[!th]
	\caption{Experimental evaluation on the LRW test set and comparison with recent methods from the literature. Results are sorted by computational complexity. ``FLOPs'' refers to Floating Point OPerations, ``(KD)" indicates that the model was trained using knowledge distillation methods. Models proposed in this work are highlighted.}
	\label{tab:results}
	\centering
	\begin{tabular}{l c c c}
		\hline
		\textbf{Method (Models used)} & \textbf{FLOPs ($\times 10^9$)} & \textbf{Parameters ($\times 10^6$)} & \textbf{Accuracy $\uparrow$ (\%)} \\
		\hline
		ShuffleNet v2 (1$\times$) + MS-TCN (KD) \cite{ma2021towards} & 2.23 & 28.8 & 85.5 \\
		ResNet-18 + MS-TCN \cite{martinez2020lipreading} & 10.31 & 36.4 & 85.3 \\
		MobiVSR-1 \cite{shrivastava2019mobivsr} & 11.0 & 4.50 & 72.2 \\
		ResNet + DC-TCN \cite{ma2021lip} & 10.64 & 52.54 & 88.36 \\
		ResNet + 3$\times$Bi-GRU \cite{feng2021efficient} & 10.54 & 59.5 & 88.4 \\
		ResNet + 2$\times$Bi-LSTM \cite{ivanko2022visual} & 10.24 & 50.07 & 88.7 \\
		DenseNet + 3$\times$Bi-GRU \cite{yang2019lrw} & 26.12 & 14.31 & 83.0\\
		
		\hline
		\bfseries ResNet (Ghost module) + MS-TCN (Ghost module) & 3.60 & 16.73 & 86.67 \\
		\bfseries ResNet (Ghost module) + MS-TCN & 4.13 & 28.02 & 87.69 \\
		
		\bfseries ResNet (Ghost module) + DC-TCN (Ghost module) & 3.85 & 29.48 & 87.58 \\
		\bfseries ResNet (Ghost module) + DC-TCN & 4.49 & 44.21 & 88.62 \\
		
		\bfseries ResNet (GhostV2 module) + MS-TCN (Ghost module) & 5.42 & 27.78 & 87.39 \\
		\bfseries ResNet (GhostV2 module) + MS-TCN & 5.94 & 39.07 & 87.51 \\
		
		\bfseries ResNet (GhostV2 module) + DC-TCN (Ghost module) & 5.67 & 40.53 & 87.82 \\
		\bfseries ResNet (GhostV2 module) + DC-TCN & 6.30 & 55.26 & 88.49 \\ 
		
		\bfseries ResNet-18 + MS-TCN (Ghost module) & 9.76 & 25.06 & 87.87 \\
		\bfseries ResNet-18 + DC-TCN (Ghost module) & 10.01 & 37.81 & 89.1 \\
		\hline
	\end{tabular}
\end{table*}

Our experimental evaluation showcases that utilizing the Ghost modules on each component of the architecture (feature extraction or sequence model), bestows a noticeable improvement in computation requirements, since the cheap operations in the Ghost module are much more lightweight compared to the regular convolutions in the original networks.
In addition, we also gain significant savings in model sizes by lowering parameter counts leading to more compact final models, allowing for applications in a broader range of devices as network size is essential for energy savings due to memory access costs.

Simultaneously, a minor accuracy drop occurs, arguably due to the reduced representation capabilities of the Ghost module, which is a drawback also mentioned in \cite{tang2022ghostnetv2}.
Nevertheless, the residual convolutional network \cite{he2016deep} equipped with Ghost modules still performs rather well, being highly competitive with larger networks, and it surpasses other works while being more lightweight in terms of both size and computation.
Notably, employing GhostV2 modules in the residual architecture causes an increase in model parameters, due to the design of DFC attention which uses two additional convolution layers (see Equation \ref{eq:dfc} in Section \ref{sub:ghost}).
When using this module, since we replace both standard convolutions within each residual block in the original network, the total number of parameters increases.
However, this added amount is offset when combined with a TCN variant that also uses the Ghost module as its building block, so the overall parameter count drops and is still lower than the original network.

Interestingly, the GhostV2 module that includes the DFC attention provides a minor accuracy improvement only in cases where both components utilize Ghost modules, indicating that the DFC attention is better utilized in a more resource-constrained network and this component can be used to recover some performance in such applications.
This added accuracy can also be explained by the larger network size, for example, when using the MS-TCN \cite{martinez2020lipreading} with Ghost modules, employing the GhostV2 module in the feature extraction network surpasses the performance of the original Ghost module ($87.39\%$ vs $86.67\%$ accuracy), but requires an additional $11.05$ million parameters and $1.82$ GFLOPs.
We note a similar observation when using the densely-connected (DC-TCN) \cite{ma2021lip} sequence model.

A more evident benefit of these modules is the considerable reduction in FLOP count, which can be more preferable than parameter savings in some scenarios (e.g., hardware with adequate memory but a low-power processing unit).
Depending on the available resources of a device, a model where only one component utilizes Ghost modules can be used to suit the application.
Table~\ref{tab:params_flops_ghost} shows a more comprehensive comparison of hardware requirements per network component when using the Ghost modules.

In the case of using Ghost modules in the temporal convolution network variants, the already lightweight TCNs are made even more compact by further reducing their size and FLOPs.
More concretely, when replacing the standard convolution layers with Ghost modules in the multi-scale model, we notice significant reductions in FLOPs ($47.3\%$), while the parameter count drops by about $44.8\%$.
Similarly, in the densely-connected TCN variant, using Ghost modules brings the total parameter count down to $~26.6$ from $~41.3$ million, a $35.6\%$ reduction, and also cuts its computation cost by around $42.8\%$, while maintaining acceptable recognition accuracy.
Finally, both of our proposed temporal convolution networks with Ghost modules outperform the ResNet-18 and MS-TCN architecture \cite{martinez2020lipreading} by $1.3\%$ and $2.1\%$ accuracy at $74.3\%$ and $47.7\%$ fewer FLOPS, respectively.

\subsection{Partial TCNs}\label{sub:partial_tcn}

We evaluate the ultra-lightweight Temporal Convolution Network variants on LRW when using our proposed Partial Temporal Block as their core component.
As mentioned previously (see Subsection~\ref{sub:partial}), we employ three architectures from the literature within our block.
The results are shown in Table~\ref{tab:partial}.

\begin{table*}[!th]
	\caption{Experimental evaluation on the LRW test set for our methods using the proposed Partial Temporal Block. In these experiments, the kernel size for all convolution operations that are not \textit{point-wise} is indicated. ``FLOPs" refers to Floating Point OPerations and parameters are measured in millions.}
	\label{tab:partial}
	\centering
	\begin{tabular}{l c c c}
		\hline
		\textbf{Method} & \textbf{FLOPs ($\times 10^9$)} & \textbf{Parameters ($\times 10^6$)} & \textbf{Accuracy $\uparrow$ (\%)} \\
		\hline
		ResNet + Partial TCN (Temporal block, k=7) & 9.59 & 22.80 & 85.29 \\
		ResNet (Ghost module) + Partial TCN (Temporal block, k=5) & 3.27 & 11.22 & 83.05 \\
		\hline
		ResNet + Partial TCN (ShuffleNet block, k=5) & 9.20 & 13.85 & 84.44 \\
		ResNet (Ghost module) + Partial TCN (ShuffleNet block, k=3) & 3.05 & 5.50 & 81.93 \\
		\hline
		ResNet + Partial TCN (FasterNet block, k=3) & 9.36 & 20.56 & 87.03 \\
		ResNet (Ghost module) + Partial TCN (FasterNet block, k=3) & 3.20 & 12.23 & 86.48 \\
		\hline
	\end{tabular}
\end{table*}

\subsection{Ablation Studies}\label{sub:ablation}

We perform an ablation analysis experimenting with the ratio used in the partial temporal block within the TCN-based sequence models, see Table~\ref{tab:ablation_ratio}.
This parameter controls the balance between the channels of each computation branch when splitting the input feature map (as shown in Figure~\ref{fig:partial_block}).
In this experiment, we use the FasterNet \cite{chen2023run} formulation (Figure~\ref{fig:fasternet_block}) as the core of our Partial Temporal Block, since it outperforms the other two methods.
In this setup, one branch has no calculations, and therefore, no overhead, meaning that the ratio effectively controls the amount of calculations per block; a higher ratio provides more channels to the branch with the resource-intensive computations, increasing overall performance at the cost of resources and vice-versa.
For feature extraction, we employ two CNNS: the standard $18$-layer residual model \cite{he2016deep} and a lightweight version with Ghost modules.
We train all models with the procedure mentioned in Subsection~\ref{sub:training}.

\begin{table*}[!th]
	\caption{Ablation analysis on the channel ratio in the partial block. Evaluation is performed in the LRW test set. ``FLOPs" refers to Floating Point OPerations ($\times 10^9$). Parameters are shown in millions ($\times 10^6$).}
	\label{tab:ablation_ratio}
	\centering
	\begin{tabular}{l c c c c}
		\hline
		\textbf{Method} & \textbf{ratio} & \textbf{FLOPs} & \textbf{Param.} & \textbf{Accuracy $\uparrow$ (\%)} \\
		\hline
		\multirow{3}{*}{ResNet + Partial TCN (FasterNet block)} & 0.25 & 9.30 & 18.9 & 85.21 \\
		 & 0.5  & 9.32 & 19.5 & 85.37 \\
		 & 0.75 & 9.36 & 20.5 & 87.03 \\
		\hline
		\multirow{3}{*}{ResNet (Ghost module) + Partial TCN (FasterNet block)} & 0.25 & 3.14 & 10.6 & 82.30 \\
		 & 0.5  & 3.16 & 11.2 & 85.40 \\
		 & 0.75 & 3.20 & 12.2 & 86.48 \\
		\hline
	\end{tabular}
\end{table*}

Using a higher ratio, as one would expect, leads to greater overall recognition accuracy, since, after splitting, the branch that performs calculations receives a larger volume and operates on a higher percentage of the input, exploiting information from more channels.
This is accompanied by a slightly higher FLOP and parameter count of the TCN-based models, which is not significant, especially when using the Ghost module, which significantly shrinks the overall costs.
Switching the ratio from $0.25$ to $0.75$ only adds $0.05$ GFLOPs and $1.6$ million parameters while raising accuracy by $1.82\%$ and up to $4.18\%$, depending on feature extraction model.
The higher ratio ($0.75$) allows the CNN with Ghost modules to achieve large accuracy gains, surpassing several networks that are much more expensive.

We also perform an additional experiment, where we increase the kernel size of the convolutions in each block, in order to provide the network with a larger effective receptive field and tabulate the results in Table~\ref{tab:ablation_kernel}.
For this experiment, we evaluate the Temporal and ShuffleNet \cite{ma2018shufflenet} architectures in our block and set the ratio to $0.75$ as it offers the best performance for a negligible impact in computation overhead.
Same as before, we keep the previous training settings.

\begin{table*}[!th]
	\caption{Ablation analysis on the kernel size used in the branch that performs operations in the partial block. Evaluation is performed in the LRW test set. ``FLOPs'' refers to Floating Point OPerations ($\times 10^9$). Parameters are shown in millions ($\times 10^6$).}
	\label{tab:ablation_kernel}
	\centering
	\begin{tabular}{l @{} c c c c}
		\hline
		\textbf{Method} & \textbf{kernel size} & \textbf{FLOPs} & \textbf{Param.} & \textbf{Acc. $\uparrow$ (\%)} \\
		\hline
		\multirow{4}{*}{ResNet + Partial TCN (Temporal block)} & 3 & 9.30 & 16.31 & 82.75 \\
		 & 5 & 9.43 & 19.55 & 83.78 \\
		 & 7 & 9.59 & 22.80 & 85.29 \\
		 & 9 & 9.80 & 26.04 & 84.10 \\
		\hline
		\multirow{4}{*}{ResNet (Ghost module) + Partial TCN (Temporal block)} & 3 & 3.14 &  7.98 & 81.19 \\
		 & 5 & 3.27 & 11.22 & 83.05 \\
		 & 7 & 3.44 & 14.46 & 82.64 \\
		 & 9 & 3.64 & 17.71 & 83.07 \\
		\hline
		\multirow{4}{*}{ResNet + Partial TCN (ShuffleNet block)} & 3 & 9.20 & 13.84 & 83.65 \\ 
		 & 5 & 9.20 & 13.85 & 84.44 \\
		 & 7 & 9.20 & 13.86 & 84.13 \\
		 & 9 & 9.20 & 13.87 & 83.37 \\
		\hline
		\multirow{4}{*}{ResNet (Ghost module) + Partial TCN (ShuffleNet block)} & 3 & 3.05 & 5.50 & 81.93 \\
		 & 5 & 3.05 & 5.52 & 81.92 \\
		 & 7 & 3.05 & 5.53 & 81.57 \\
		 & 9 & 3.05 & 5.54 & 81.68 \\
		\hline
	\end{tabular}
\end{table*}

Generally, using a larger kernel size improves recognition accuracy while slightly raising the overhead due to the amount of calculations required by the larger kernel.
We note however, that this does not apply to all cases, for instance, when using the ShuffleNet block, a larger kernel size than $5$ (e.g., $7, 9$) does not improve accuracy and in fact, hampers performance when the residual network with Ghost modules is used.
For a more clear overview of the complexity that each component adds to the overall measurements, the reader is referred to Section~\ref{sub:param_analysis}.

As for the TCN using the Temporal block, it scales better with a larger kernel size, improving its performance, compared to the ShuffleNet block,
however, this network's FLOPS and parameters increase at a much higher rate since it uses regular convolutions.
The same diminishing effect in accuracy gains is noticed for the largest kernel sizes.
We believe this result is caused by the the dilation amount used in the deeper layers of the TCN architecture, which causes the larger kernels to miss information from their input.
Similar to the results shown in previous tables (e.g., Table~\ref{tab:results}), when Ghost modules are used in the convolutional feature extraction network, significant reductions in computation and sizes are gained, while the final accuracy suffers slightly.

\subsection{Parameter Analysis}\label{sub:param_analysis}

In addition, we gather all measurements related to network size and complexity for all proposed architectures in this work and provide the results in Tables~\ref{tab:params_flops_ghost} and ~\ref{tab:parameter_analysis}, showcasing the efficiency gained by using Ghost modules and our proposed partial block when designing lightweight networks.

\begin{table}[!th]
	\caption{Detailed parameter analysis per component when using the Ghost modules. Proposed TCN with \textit{FasterNet} \cite{chen2023run} block also added for comparison. ``FLOPs" refers to Floating Point OPerations ($\times 10^9$), while parameters are measured in millions ($\times 10^6$).}
	\label{tab:params_flops_ghost}
	\centering
	\begin{tabular}{l c c}
		\hline
		\textbf{Model} & \textbf{FLOPs} & \textbf{Parameters} \\
		\hline
		ResNet-18 & 8.29 & 11.16 \\
		ResNet-18 (Ghost module) & 2.13 (-74.3\%) & 2.83 (-74.6\%) \\
		ResNet-18 (Ghost module + DFC) & 3.95 (-52.3\%) & 13.88 (+19.5\%) \\
		\hline
		MS-TCN & 1.12 & 25.17 \\
		MS-TCN (Ghost module) & 0.59 (-47.3\%) & 13.88 (-44.8\%) \\
		\hline
		DC-TCN & 1.47 & 41.36 \\
		DC-TCN (Ghost module) & 0.84 (-42.8\%) & 26.63 (-35.6\%) \\
		\hline
		TCN (FasterNet block, ratio=0.25) & 0.12 & 7.80 \\
		TCN (FasterNet block, ratio=0.5) & 0.15 & 8.39 \\
		TCN (FasterNet block, ratio=0.75) & 0.18 & 9.38 \\
		\hline	
	\end{tabular}
\end{table}

As mentioned previously, the convolution networks using Ghost modules achieve significant savings in both size and computation compared to the standard model, with the exception of DFC attention module which adds $~2.72$ million parameters to the original $18$-layer ResNet \cite{he2016deep} architecture.
In both cases, however, we note a measurable reduction in FLOP count, where the Ghost module requires only $2.13$ GFLOPs, a $74.3\%$ reduction from the original.
Due to the added computation, the DFC attention module is a bit more expensive at $3.95$ GFLOPs, which, nevertheless, is still less than half of the standard model.
For the sequence models, a more modest reduction in size and overhead (up to $44.8\%$ and $47.3\%$ respectively, in the case of the MS-TCN) is achieved, since the original architectures are already quite lightweight.
Due to the added complexity of the DC-TCN network, the reductions less significant than in the other architectures.

Finally, the TCN-based architectures using our proposed Partial Temporal Block are even more lightweight regardless of block design.
These variants require a fraction of resources compared to all other architectures and scale favorably with ratio as well as kernel size.
Overall, the FasterNet \cite{chen2023run} design is the superior choice for our proposed Partial Temporal Block as it maintains a very low FLOP and parameter overhead across all channel ratios, and at the highest setting ($0.75$) it outperforms several larger models as well as the other Partial TCN variants.
The ShuffleNet \cite{ma2018shufflenet} design also maintains extremely low FLOP and parameter measurements but falls behind the other designs in performance mainly due to the rather low parameter count.
When combined with the Residual network with Ghost modules, it forms a highly compact overall model at around $5.5$ million parameters that is more suitable for hardware with very low capabilities.
The Temporal block design presents a performance compromise between the two previous architectures, surpassing the ShuffleNet design, while also maintaining low FLOPs as we increase the kernel size, but this block is hindered by increasing parameter counts.

\begin{table}[!th]
	\caption{Size and complexity analysis of the TCN variants using our proposed partial block for different core components and kernel sizes. Evaluation is performed in the LRW test set. ``FLOPs" refers to Floating Point OPerations ($\times 10^9$), while parameters are measured in millions ($\times 10^6$). The ratio for the channel split used in all models in this table is set to $0.75$ as it is the most resource-intensive amount.}
	\label{tab:parameter_analysis}
	\centering
	\begin{tabular}{l c c c}
		\hline
		& \textbf{Block} & \textbf{ShuffleNet} & \textbf{Temporal} \\
		\textbf{Kernel size} & & & \\
		\hline
		\multirow{2}{*}{3} & FLOPs & 0.34 & 0.12 \\
		 & Parameters & 1.20 & 3.80 \\
		\hline
		\multirow{2}{*}{5} & FLOPs & 0.35 & 0.25 \\
		 & Parameters & 1.20 & 6.18 \\
		\hline
		\multirow{2}{*}{7} & FLOPs & 0.35 & 0.42 \\
		 & Parameters & 1.21 & 8.52 \\
		\hline
		\multirow{2}{*}{9} & FLOPs & 0.36 &  0.62 \\
		 & Parameters & 1.21 & 10.87 \\
		\hline
	\end{tabular}
\end{table}

\subsection{Limitations}\label{sub:limitations}

A current drawback of the DFC attention block lies in its design which exploits two convolutions in two directions (vertical and horizontal).
This prevents its exploitation by the temporal networks which utilize 1D convolutions, and for this reason in our models, its use is limited in the residual convolutional architecture which serves as a feature extractor.
Also, in Table~\ref{tab:results} is shown that this module does not bring improvements in all cases where it is used, for example, when the sequence model does not employ Ghost modules.
A possible explanation is that the DFC module was originally designed for images of higher dimensions ($224\times224$) and its use is sub-optimal in that architecture due to the fact that the 3D convolution and pooling block at the beginning of the overall model reduce the spatial dimensions of the feature map.
The additional down-sampling (see Section~\ref{sub:ghost}, Equation~\ref{eq:dfc}) performed by the DFC attention module of the (already low-dimension) feature map removes much of the information contained and hinders the module's ability to exploit it.
We believe that removing the pooling operations could possibly improve the overall performance, slightly increasing the computational complexity, and plan on investigating this in the future.

\section{Conclusion}\label{sec:conclusion}

In this work, we proposed taking advantage of low-cost components to develop lightweight architectures for practical visual speech recognition (VSR) applications.
Using the recently proposed Ghost modules where an amount of the channels within are calculated with cost-efficient operations, we developed low-resource models for VSR of isolated words.
We replaced the standard convolution operations with Ghost modules in the visual extraction and sequence modeling networks creating compact and efficient alternatives that showcase significantly lowered computational resource requirements.
Their reduced overhead enables a multitude of applications in several scenarios where speed of operation is critical and hardware resources are constrained.
Evaluation on the largest single word speech recognition dataset showed that our models outperform other lightweight architectures while demanding fewer computational resources measured in FLOPs.
Simultaneously, the achieved accuracy of the models is very competitive with other architectures that are much larger in terms of model size and complexity.
Moreover, we proposed a general component called "Partial Temporal block" for building ultra-lightweight sequential models intended for devices with very limited hardware capabilities, such as IoT and edge devices.
This block splits the computation path in two branches and can be customized to fit each use case according to the task and resources at hand.

Future work includes addressing the weaknesses outlined in this work, i.e., architectural tuning to take advantage of the DFC module and the larger kernel sizes.
We also intend to expand our proposed partial block's capabilities by exploring automated techniques for optimal operation selection, as well as introducing other efficient channel attention methods to increase performance.
Finally, specialized training strategies exploiting the latest augmentation and weight averaging approaches are also planned.


\bibliography{IEEEabrv,bibliography}
\bibliographystyle{IEEEtran}

\end{document}